# A Combination of Multi-Objective Genetic Algorithm and Deep Learning for Music Harmony Generation

Maryam Majidi and Rahil Mahdian Toroghi

Tehran, Iran

**Abstract.** Automatic Music Generation (AMG) has become an interesting research topic for many scientists in artificial intelligence, who are also interested in the music industry. One of the main challenges in AMG is that there is no clear objective evaluation criterion that can measure the music grammar, structural rules, and audience satisfaction. Also, original music contains different elements that should work together, such as melody, harmony, and rhythm; but in the most of previous works, AMG works only for one element (e.g., melody). Therefore, in this paper, we propose a Multi-Objective Genetic Algorithm (MO-GA) to generate polyphonic music pieces, considering grammar and listener satisfaction. In this method, we use three objective functions. The first objective function is the accuracy of the generated music piece, based on music theory; and the other two objective functions are modeled scores provided by music experts and ordinary listeners. The scoring of experts and listeners separately are modeled using Bi-directional Long Short-Term Memory (Bi-LSTM) neural networks. The proposed music generation system tries to maximize mentioned objective functions to generate a new piece of music, including melody and harmony. The results show that the proposed method can generate pleasant pieces with desired styles and lengths, along with harmonic sounds that follow the grammar.

## 1 Introduction

By growing the multimedia contents, attracting audiences is becoming more difficult day by day. In the meantime, music may be one of the popular options because of its ability to evoke emotions. Digital advances have also changed the shape of music composing. In recent years several methods have been introduced for Automatic Music Generation (AMG) using computers and algorithms. The machine, like humans, can generate music pieces using the rules and music elements. Also, it can help musicians and composers to make new ideas or to generate effective content.

AMG models can be classified into these groups: Markov model-based methods [1-7], approaches based on music rules and regulations [8-14], neural network-based models [15-22], methods based on evolutionary optimization algorithms, and population-based [23-28], and algorithms based on local search [29-30].

Most of the mentioned methods focus on generating melody and rhythm, and do not consider harmony, or generating appropriate chords. Also, most of the methods do not consider human taste or satisfaction in the generation process. Among the previous works, the study of Farzaneh and Mahdian [28], proposes an evolutionary model interacting with humans as an objective function. But their evaluations are only based on the opinions of regular listeners. Also, it doesn't consider the grammars.

In this paper, we use a multi-objective evolutionary algorithm for harmony generation. This method interacts with both regular and expert listeners. Furthermore, we use music grammars as the third objective function. So, the proposed method, in each execution tries to maximize the listener's satisfaction (regular and experts) as well as the accuracy of the generated music piece according to the music theory and grammar. In order to remove the listeners in the AMG process, and completely automizing the system, we have modeled the human scoring using two Bi-LSTM neural networks (one for regular listeners and another for expert listeners).

The rest of the paper is organized as follows:

Section 2 provides state-of-arts in AMG. Section 3 introduces the proposed AMG system. Section 4 reports the experimental results, and section 5 concludes the work.

## 2  Related Works

In this section, we review the literature for AMG. Some of the early works of melody generation used random models. In 2010, Davismoon and Eccles were among the first researchers to introduce melody generation as a hybrid optimization problem with a Markov model integrated into the objective function [9]. Herremans and Chuan take a different approach, inspired by linguistics. They used neural networks to evaluate the ability of semantic vector model patterns (word2vec) to record music text and semantic similarity [21].

In recent years, some researchers have shown the effectiveness of using techniques such as optimization and deep learning over earlier methods such as Markov models. More sophisticated deep learning models such as recursive neural networks have become popular. This trend is partly due to the fact that such models can learn complex relationships between notes with respect to existing components. Some of these models make it possible to create music pieces by repeating patterns and structural concepts. Franklin created a recurrent neural network with Long Short-Term Memory (LSTM) that generates melody or monophonic music pieces [31].

Harmony or polyphonic music is a process of combining individual voices that are analyzed by hearing them simultaneously. The science of harmony often refers to the vertical aspect of music and is distinct from horizontal and melodic motion. The purpose of using harmony is to accompany the melodies by considering the relevant rules that lead to the creation of polyphonic music. Among the works done in the field of harmony, we can mention the synchronization of melodies using the evolutionary optimization algorithm [32]. Chorale generation is one of the most popular works of music generation in terms of harmony and produces very structured music. The most common form is the generation of three sounds, which are used to harmonize a certain melody. The Bach-in-a-Box system provided by McIntyre harmonizes a melody created by the user [33].

Nakamura et al created a system that automatically generates sound effects and background music for short videos. The melody, harmony, and rhythm of each scene are created by considering its moods, its intensity, and weakness [34]. Tuohy and Potter developed a genetic algorithm that produces playable guitar music by minimizing hand and finger movements [35].
One of the recent AMG approaches is the use of Generative Adversarial Networks (GANs) GANs as well as reinforcement learning methods. In 2017, Yang et al proposed a system called MidiNet, which in two separate experiments generated melody and melody with harmony by combining the GAN and Convolutional Neural Networks (CNN) [36].

## 3   The Proposed Method

At the first stage a genetic algorithm (GA1) is used to generate a wide range of music pieces (melodies with chords). The objective function of this GA considers the rhythms and similarity between generated pieces and a standard database of polyphonic human-made pieces. So, after the first stage, we have a collection of optimized music pieces. Then two groups of expert listeners and regular listeners provide scores between 0 and 100 for each piece in the generated collection. The scoring of each group is then modeled by a Bi-LSTM neural network. Then we run another genetic algorithm (GA2) by adding the trained networks to the objective function. In this way, the multi-objective algorithm considers both the rules and the satisfaction of human listeners in a generation. The final music generation system is shown in Fig. 1.

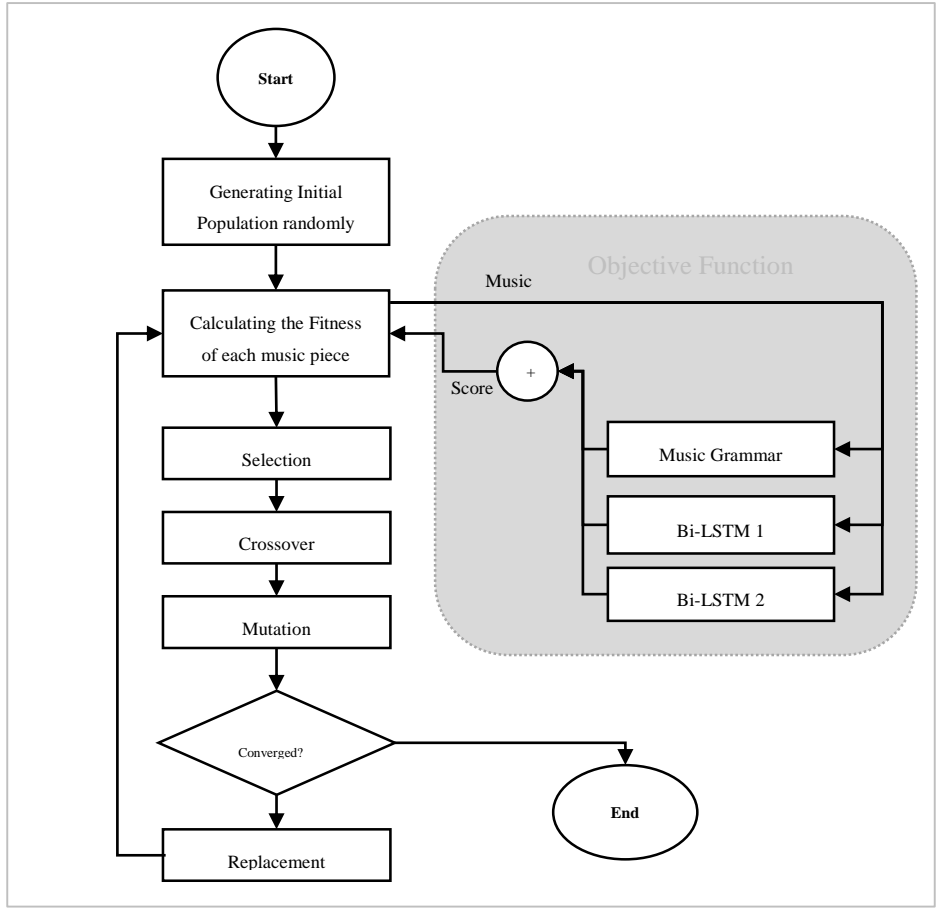

**Fig. 1.** The proposed music generation system.

### 3.1 Chromosome Structure

Each chromosome in GA is actually a music piece. So, GA tries to optimize the music piece based on its objective functions. To generate chromosomes, we convert each music piece (including melody and chords) from ABC notation into a piano matrix.

The piano matrix contains 88 rows for 88 keys on a piano, and optional columns as the number of time intervals. Each time interval is equivalent to 1/64 of the original time unit of the music.

This matrix only contains ones and zeros. One means the presence of sound at a certain key on the piano, and zero means silence. So, according to the letter of each note in ABC notation and its octave, we specify its corresponding row and numbering the matrix elements based on the length of each note.

## 3.2 Genetic Algorithm for Data Collection (GA1)

Using GA1, we generate a collection of music pieces (melody with chords) considering the music rules. At each iteration of GA1, the best chromosomes in the population are selected based on minimum violation of the rules and the maximum similarity to a human-made polyphonic music database.

So, we use a database as a reference and compute the similarities of Bi-gram, Tri-gram, and 4-gram in the generated pieces. In addition, there are costs for violating the rules so that the transmission of the notes to each other is musically sensible and a pleasant frequency interval is created. Therefore, we maximize the probability of each note occurring after another note. For chords that match the melodies, we follow the rules of chord writing and also extract repetitive combinations of simultaneous sounds from the database. In addition, to break the rhythm, a cost is considered in the objective function. Then, the fitness function is calculated from Equation 1 which should be maximized.

$$Objective\ Function = Score + \frac{e}{e+Cost} \qquad (1)$$

Where *Cost* is the number of rhythm and harmony violations. The constant $e$ is a small value to prevent division by zero, and *Score* is the similarity value between the chromosome (generated music piece) and the human-made database:

$$Score = \frac{N_2 + 10N_3 + 100N_4 + S_2 + 10S_3 + 100S_4}{ML} \qquad (2)$$

Where $N_2$, $N_3$, and $N_4$ are the numbers of repeated Bi-grams, Tri-grams, and 4-grams in the chromosome and the database respectively (similar rows). $S_2$, $S_3$, and $S_4$ are the simultaneous combinations of two-voice, three-voice, and four-voice repetitive between the chromosome and the database (similar columns). $M$ is the number of notes in the human-made database, and $L$ is the number of notes in the chromosome.

To consider the style of the database and transfer it to the outputs, we calculate the probability of occurrence of each note in that style, and based on these probabilities, random generation by genetics occurs. The character probability can be calculated as follows:

$$P_{n_i} = \frac{n_i}{n_{total}} \qquad (3)$$

Where $n_i$ is the number of occurrences of each character in the entire database, and $n_{total}$ is the total number of characters in the database.

At each iteration, crossover and mutation are applied after selecting the most competent chromosomes. The pseudocode for the crossover is as follows,

```
START
        PROGRAM Crossover Operator
        READ best1 and best2
        Half a child is equal to half a best1
        Half a child is equals to half a best2
END
```

Where *best1* and *best2* are the selected chromosomes, and the *child* is a new chromosome after crossover.

The pseudocode for mutation is as follows,

```
START
   PROGRAM Mutation Operator
   READ child
   IF Mutation Rate is equal to 0.1
      FOR 1 through Number of channels
         FOR 1 through Number of child sizes
            IF the random number is less than the Mutation rate and child is
            not equal to zero
            SET the child equal to zero
            END IF
            IF the random number is less than the Mutation rate and child is equal to
            zero
            SET the child equal to Number of channels
            END IF
         END FOR
      END FOR
   END IF
END
```

Where the *channel* is the number of simultaneous sounds.

### 3.3  Bi-LSTM-based Evaluating Models

Each generated music piece from the previous stage is heard by two groups of listeners: experts and regular listeners. from the average score that people provide, the music piece fitness is determined.

To simulate the way of scoring for each group of listeners, a Bi-LSTM neural network is used. Since all the musical notes are interconnected and their connection is maintained to the end, the Bi-directional LSTM recursive neural network is used, which has a memory and examines the logical connection of the notes from the beginning to the end, and from the end to the beginning. These networks can provide the output score with the least error by solving a regression problem from input samples that have long-term dependencies. The output range is between 0 and 100. The networks receive a chromosome as a piano matrix which is a sequence and provide scores that show the fitness of each music piece. After training, networks can be used as criteria for evaluating musical sequences and their satisfaction. The architecture of the LSTM network is shown in Fig. 2.

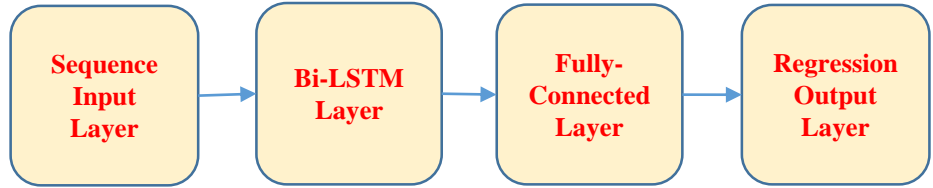

**Fig. 2.** The architecture of the LSTM network.

### 3.4 Proposed AMG system (GA2)

As can be seen in Figure 3, GA2 for the AMG system has the same process as the GA1, except that, the fitness function, by adding human, has several objectives: Music rules, Bi-LSTM trained neural network model with opinions of music experts, and the trained Bi-LSTM neural network model with opinions of regular listeners. We calculate the fitness function of the final automatic music generation system in the following equation,

$$Fitness\ Function = w_1 X_1 + w_2 X_2 + w_3 X_3 \qquad (4)$$

Where $X_1$, $X_2$, and $X_3$ are the musical grammar score, the average score of the expert listeners, and the average score of the regular listeners, respectively, provided by trained neural network models.
$w$ constants are the weights for each objective function to normalize all the objectives.

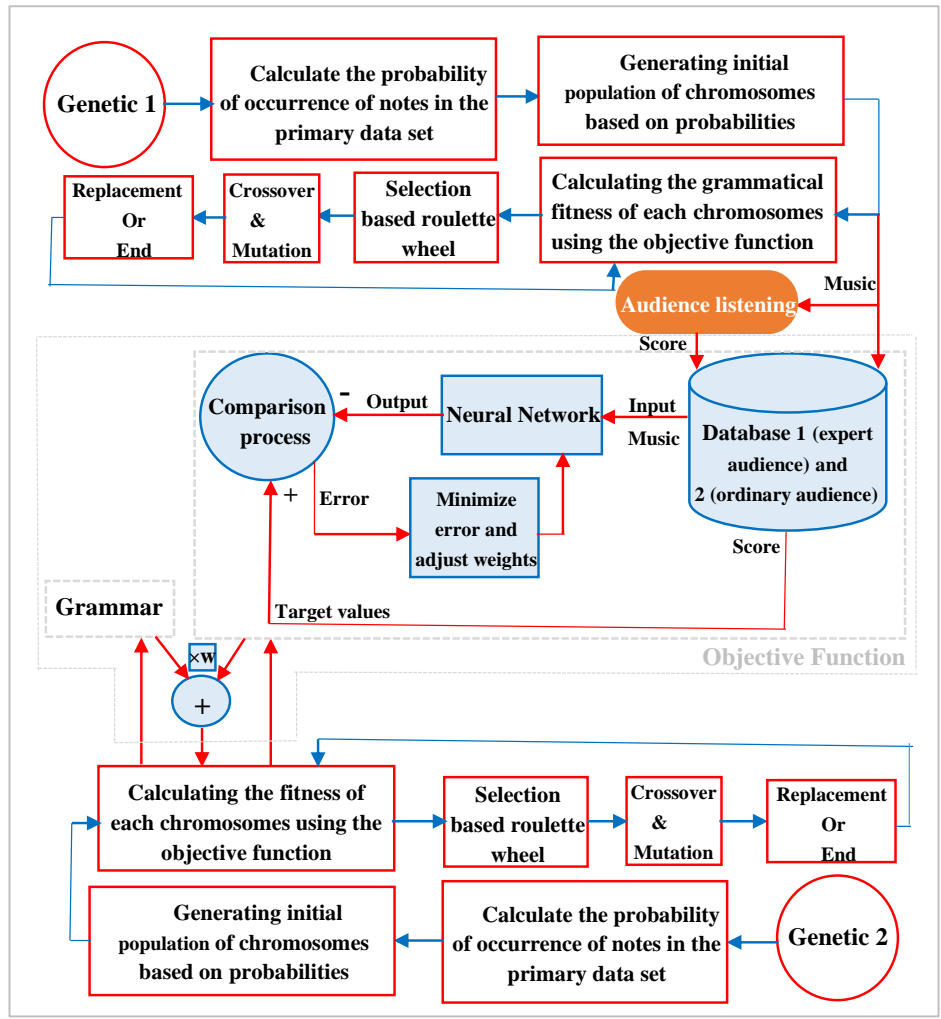

**Fig. 3.** The flowchart of the Music Harmony Generation System.

## 4     Experimental Results

The proposed system is implemented in MATLAB R2018b, and used the Steirar database,[1] which contains 235 polyphonic music pieces in ABC notation. Figure 4 shows the probability of occurrence of notes in this database.

---

[1] http://abcnotation.com/tunes

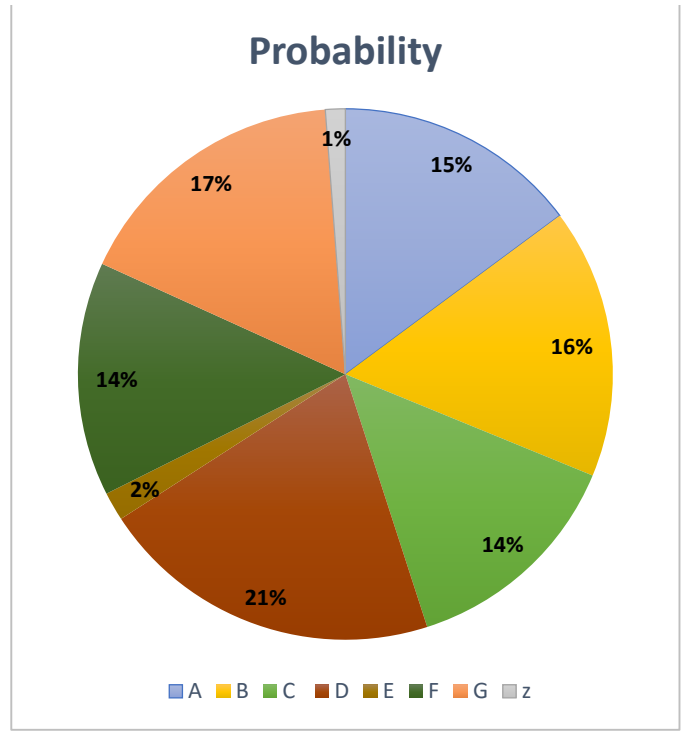

**Fig. 4.** Note's probability of occurrence in Steirar database. Musical notes are represented in the ABC notation standard.

The parameters of each algorithm can be seen in table 1:

**Table 1.** Parameters Settings

| Parameter | GA 1 | GA 2 |
|---|---|---|
| **Number of Iterations** | 3600 | 3600 |
| **Population Size** | 15 | 15 |
| **Crossover rate** | 0.5 | 0.5 |
| **Mutation rate** | 0.1 | 0.1 |
| **Objective Function** | Grammar | Grammar & Human |

In Bi-LSTM neural networks, four layers are used, and:

- The first layer in networks is actually the input layer that takes the matrix of notes (piano matrix).
- The second layer is the Bi-LSTM layer, which has 50 neurons in the hidden layer.
- The third layer is a Fully Connected layer with one output.
- The last layer is a regression layer with a cost function of the Root of the Mean Square Error (RMSE). This layer provides the score of the input music piece.
- The number of learning epochs is set to 5,000.

After executing the GA2, we can compare the results of the two steps. Figure 5 shows the time and speed of generation to generate different music pieces with different lengths and 2 simultaneous sounds. It is observed that the GA2 is faster than the GA1.

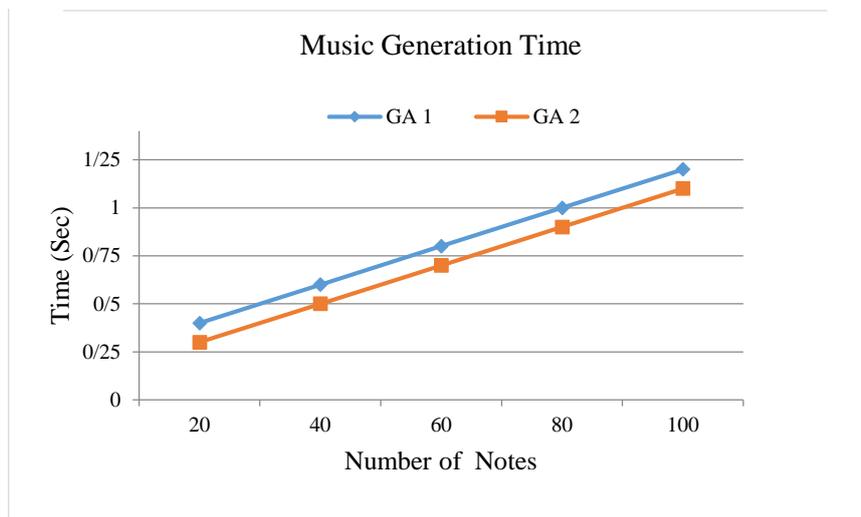

**Fig. 5.** The execution time at GA1, and GA2 vs. the number of notes.

We also gave 5 outputs of GA1, and 5 outputs of GA2 expert and regular listeners, and asked them to give a score between 0 and 100 to the outputs. Figures 6 and 7 show their average scores. Fig. 8 shows a sample output of the proposed AMG system.

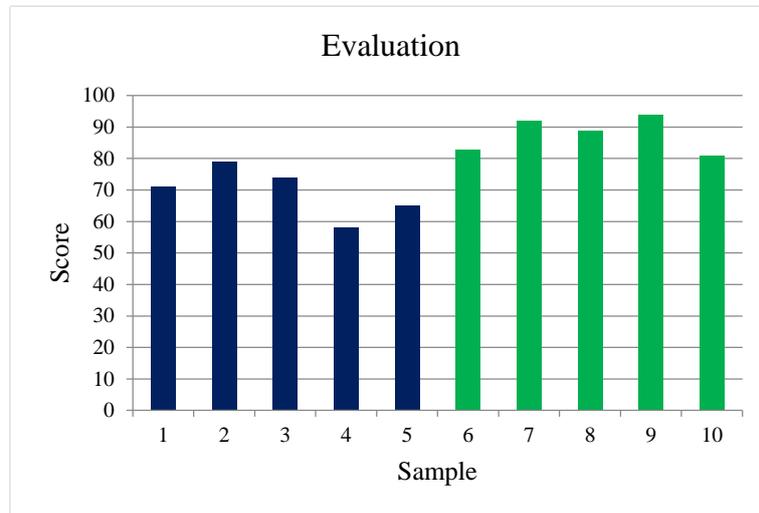

**Fig. 6.** Evaluation to ensure compliance with music rules, scored by 15 expert audiences. Average scores vs. Music samples.

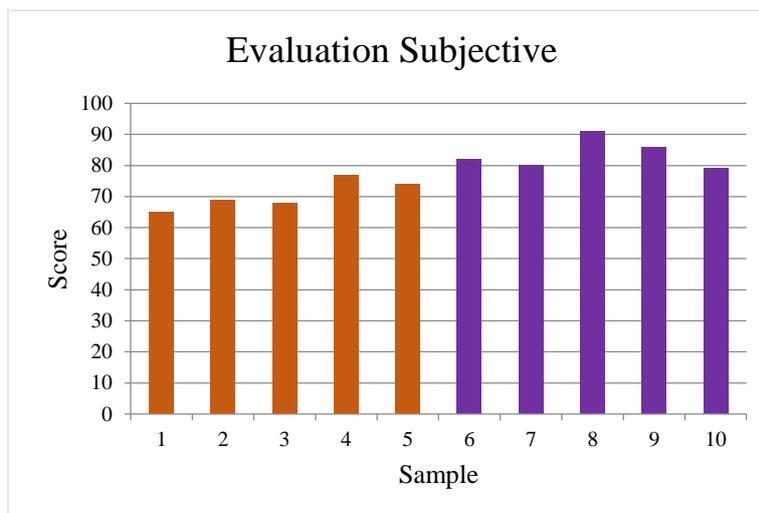

**Fig. 7.** Subjective evaluation of the music samples. Beauty and pleasurability of the music, scored by 15 audiences. Average scores vs. Music samples.

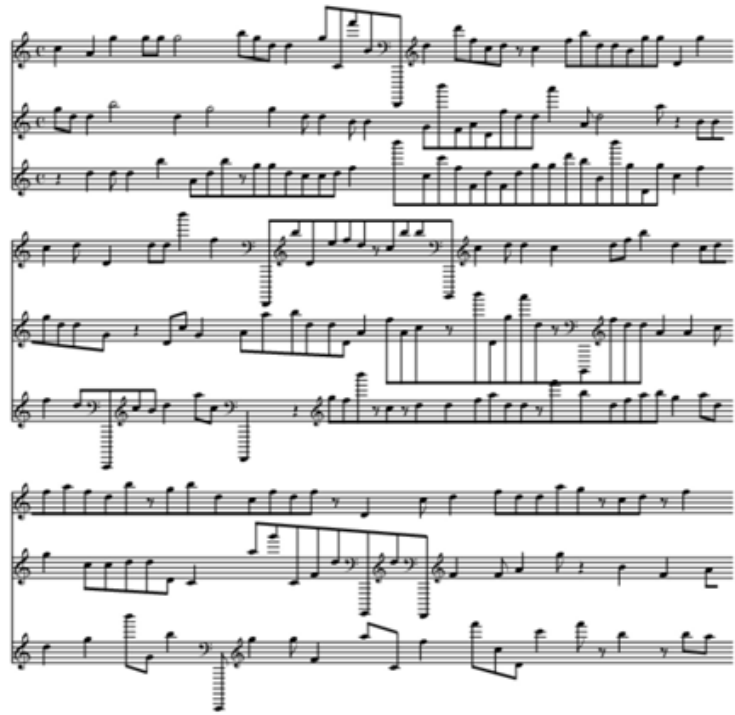

**Fig. 8.** Sample output of the proposed music generation system.

## 5      Conclusion

In this paper, we proposed a method for the automatic generation of polyphonic music. In the proposed method, we first generated a collection of polyphonic music pieces using a genetic algorithm and a database of human-made pieces with the objective function. We then stored the range of different outputs generated in this section for scoring expert and normal listeners and trained the two Bi-LSTM artificial neural networks using the opinions of these two groups of listeners. Finally, we used another genetic algorithm and set the music rules as well as trained models as separate objectives for the algorithm. In this way, living (human) and non-living (musical grammar) evaluators were simultaneously used in the automatic music generation. The results showed that the use of neural networks and other objective functions together accelerates the convergence. In a subjective evaluation, it was observed that more desirable samples of music could be achieved, which were able to obtain high scores from the audiences.